\pdfoutput=1

\documentclass[11pt]{article}

\usepackage{acl}

\usepackage{times}
\usepackage{latexsym}

\usepackage[T1]{fontenc}

\usepackage[utf8]{inputenc}
\usepackage{hyperref}       
\usepackage{url}            
\usepackage{booktabs}       
\usepackage{amsfonts}       
\usepackage{nicefrac}       
\usepackage{microtype}      
\usepackage{xfrac}
\usepackage{amssymb}
\usepackage{enumitem}
\usepackage{amsmath}
\usepackage{amsthm}
\usepackage{mathtools}

\DeclareMathOperator*{\softmax}{\mathsf{softmax}}

\DeclareMathOperator*{\argmin}{arg\,min}

\usepackage{marvosym}
\usepackage{microtype}

\title{$\infty$-former: Infinite Memory Transformer}

\author{Pedro Henrique Martins\textsuperscript{\Neptune} \quad
        Zita Marinho\textsuperscript{\Moon\Scorpio} \quad
        Andr\'e F.~T. Martins\textsuperscript{\Neptune\Pluto\Saturn} \\
\textsuperscript{\Neptune}Instituto de Telecomunica\c{c}\~oes~
\textsuperscript{\Moon}DeepMind~
\textsuperscript{\Scorpio}Institute of Systems and Robotics\\
\textsuperscript{\Pluto}LUMLIS (Lisbon ELLIS Unit), Instituto Superior T\'ecnico~
\textsuperscript{\Saturn}Unbabel\\
Lisbon, Portugal\\
\href{mailto:pedrohenriqueamartins@tecnico.ulisboa.pt}{\tt pedrohenriqueamartins@tecnico.ulisboa.pt},\\
\href{mailto:zmarinho@google.com}{\tt zmarinho@google.com}, \quad
\href{mailto:andre.martins@unbabel.com}{\tt andre.t.martins@tecnico.ulisboa.pt}.
}

\begin{document}
\maketitle

\begin{abstract}
Transformers are unable to model long-term memories effectively, since the amount of computation they need to perform grows with the context length. 
While variations of efficient transformers have been proposed, they all have a finite memory capacity and are forced to drop old information. 
In this paper, we propose the  \textit{$\infty$-former}, which extends the vanilla transformer with an \emph{unbounded} long-term memory. 
By making use of a continuous-space attention mechanism to attend over the long-term memory, the $\infty$-former's attention complexity becomes independent of the context length, trading off memory length with precision.
In order to control where precision is more important, $\infty$-former maintains ``sticky memories,'' being able to model arbitrarily long contexts while keeping the computation budget fixed.
Experiments on a synthetic sorting task, language modeling, and document grounded dialogue generation demonstrate the $\infty$-former's ability to retain information from long sequences.\footnote{The code is available at \url{https://github.com/deep-spin/infinite-former}.}
\end{abstract}

\section{Introduction}
When reading or writing a document, it is important to keep in memory the information previously read or written. Humans have a remarkable ability to remember long-term context, keeping in memory the relevant details \citep{carroll2007psychology,kuhbandner2020long}.
Recently, transformer-based language models have achieved impressive results by increasing the context size \cite{radford2018improving,radford2019language,dai2019transformer,rae2019compressive,brown2020language}. However, while humans process information sequentially, updating their memories continuously, and recurrent neural networks (RNNs) update a single memory vector during time, transformers do not -- they exhaustively query every representation associated to the past events.
Thus, the amount of computation they need to perform grows with the length of the context, and, consequently, transformers have computational limitations about how much information can fit into memory. For example, a vanilla transformer requires quadratic time to process an input sequence and linear time to attend to the context when generating every new word. 

Several variations have been proposed to address this problem \citep{tay2020efficient}. Some propose using sparse attention mechanisms, either with data-dependent patterns \citep{kitaev2020reformer,vyas2020fast,tay2020sparse,roy2021efficient,wang2021cluster} or data-independent patterns \citep{child2019generating,beltagy2020longformer,zaheer2020big}, reducing the self-attention complexity \citep{katharopoulos2020transformers,choromanski2020rethinking,peng2021random,jaegle2021perceiver}, and caching past representations in a memory \citep{dai2019transformer,rae2019compressive}. 
These models are able to reduce the attention complexity, and, consequently, to scale up to longer contexts. However, as their complexity still depends on the context length, they cannot deal with unbounded context.

In this paper, we propose the \textbf{\mbox{$\infty$-former}} (\emph{infinite former}; Fig.~\ref{fig:infinite_transformer}): a transformer model extended with an unbounded long-term memory (LTM), which allows the model to attend to arbitrarily long contexts. 
The key for making the LTM unbounded is a continuous-space attention framework \citep{martins2020sparse} which trades off the number of information units that fit into memory (basis functions) with the granularity of their representations. 
In this framework, the input sequence is represented as a \textbf{continuous signal}, expressed as a linear combination of $N$ radial basis functions (RBFs). By doing this, the \mbox{$\infty$-former}'s attention complexity is ${\mathcal{O}(L^2 + L\times N)}$ while the vanilla transformer's is $ \mathcal{O}(L \times (L+L_{\mathrm{LTM}}))$, where $L$ and $L_{\mathrm{LTM}}$ correspond to the transformer input size and the long-term memory length, respectively (details in \S\ref{sec:complexity}).
Thus, this representation comes with two significant advantages: (i) the context can be represented using a number of basis functions $N$ smaller than the number of tokens, reducing the attention computational cost; and (ii) $N$ can be \textit{fixed}, making it possible to represent unbounded context in memory, as described in \S\ref{sec:unbounded_memory} and Fig.~\ref{fig:unbounded_memory}, without increasing its attention complexity. 
The price, of course, is a loss in resolution: using a smaller number of basis functions leads to lower precision when representing the input sequence as a continuous signal, as shown in Fig. \ref{fig:sorting_results}.

To mitigate the problem of losing resolution, we introduce the concept of ``sticky memories'' (\S\ref{sec:sticky_mem}), in which we attribute a larger space in the LTM's signal to regions of the memory more frequently accessed. This creates a notion of ``permanence'' in the LTM, allowing the model to better capture long contexts without losing the relevant information,  which takes inspiration from long-term potentiation and plasticity in the brain \citep{mills2014cognitive,bamji2005cadherins}.

To sum up, our contributions are the following:
 \begin{itemize}
     \item We propose the \mbox{$\infty$-former}, in which we extend the transformer model with a continuous long-term memory (\S \ref{sec:ltm}). Since the attention computational complexity is independent of the context length, the \mbox{$\infty$-former} is able to model long contexts.
     \item We propose a procedure that allows the model to keep unbounded context in memory (\S \ref{sec:unbounded_memory}).
    \item We introduce sticky memories, a procedure that enforces the persistence of important information in the LTM (\S \ref{sec:sticky_mem}). 
     \item We perform empirical comparisons in a synthetic task (\S \ref{sec:sorting}), which considers increasingly long sequences, in language modeling (\S \ref{sec:gpt2}), and in document grounded dialogue generation (\S \ref{sec:dgg}). These experiments show the benefits of using an unbounded memory. 
 \end{itemize}

\section{Background}

\subsection{Transformer}
\label{sec:transformer}
A transformer \cite{vaswani2017attention} is composed of several layers, which encompass a multi-head self-attention layer followed by a feed-forward layer, along with residual connections \cite{he2016deep} and layer normalization \cite{ba2016layer}.

Let us denote the input sequence as ${X =[x_1,\dots,x_L]  \in \mathbb{R}^{L \times e}} $, where $L$ is the input size and $e$ is the embedding size of the attention layer.
The queries $Q$, keys $K$, and values $V$, to be used in the multi-head self-attention computation are obtained by linearly projecting the input, or the output of the previous layer, $X$, for each attention head $h$: 
\begin{align}
    \!\!\! Q_h = X_h W^{Q_h},\; K_h = X_h W^{K_h},\; & V_h = X_h W^{V_h},\!
\end{align}
where ${W^{Q_h}, W^{K_h}, W^{V_h} \in \mathbb{R}^{d\times d}}$ are learnable projection matrices, $d=\sfrac{e}{H}$, and $H$ is the number of heads. 
Then, the context representation ${Z_h \in \mathbb{R}^{L\times d}}$, that corresponds to each attention head $h$, is obtained as:
\begin{equation}\label{transformer_dot}
    Z_h = \softmax \left( \dfrac{Q_h K_h ^\top}{\sqrt{d}} \right)  V_h,
\end{equation}
where the softmax is performed row-wise.
The head context representations are concatenated to obtain the final context representation ${Z \in \mathbb{R}^{L\times e}}$:
\begin{equation}\label{eq:context_representation}
    Z = [Z_1,\dots,Z_{H}]W^R,
\end{equation}
where ${W^R \in \mathbb{R}^{e \times e}}$ is another projection matrix that aggregates all head's representations.

\subsection{Continuous Attention}

Continuous attention mechanisms \citep{martins2020sparse} have been proposed to handle arbitrary continuous signals, where the attention probability mass function over words is replaced by a probability \textit{density} over a signal. This allows time intervals or compact segments to be selected. 

To perform continuous attention, the first step is to transform the discrete text sequence represented by ${X \in \mathbb{R}^{L\times e}}$ into a continuous signal. 
This is done by expressing it as a linear combination of basis functions.
To do so, each $x_i$, with ${i \in \{1,\dots,L\}}$, is first associated with a position in an interval, ${t_i \in [0,1]}$, \textit{e.g.}, by setting $t_i = i/L$. Then, 
we obtain a continuous-space representation ${\bar{X}(t) \in \mathbb{R}^e}$, for any $t \in [0,1]$ as:

\begin{equation}\label{eq:continuous_embeddings}
    \bar{X}(t)=B^\top \psi(t),
\end{equation}
where ${\psi(t) \in \mathbb{R}^N}$ is a vector of $N$ RBFs, e.g., $\psi_j(t) = \mathcal{N}(t; \mu_j, \sigma_j)$, with $\mu_j \in [0,1]$, and ${B \in \mathbb{R}^{N\times e}}$ is a coefficient matrix. $B$ is obtained with 
multivariate ridge regression 
\citep{brown1980adaptive} so that ${\bar{X}(t_i) \approx x_i}$ for each $i \in [L]$, which leads to the closed form (see App.~\ref{app:reg} for details):
\begin{align}\label{eq:B}
    B^\top = X^\top F^\top (FF^\top + \lambda I)^{-1} = X^\top G,
\end{align}
where ${F=[\psi(t_1),\dots,\psi(t_L)]\in \mathbb{R}^{ N\times L}}$ packs the basis vectors for the $L$ locations. As ${G \in \mathbb{R}^{L\times N}}$ only depends of $F$, it can be computed offline.

Having converted the input sequence into a continuous signal $\bar{X}(t)$, the second step is to attend over this signal. To do so, instead of having a discrete probability distribution over the input sequence as in standard attention mechanisms (like in Eq. \ref{transformer_dot}), we have a probability density $p$, which can be a Gaussian ${\mathcal{N}(t; \mu, \sigma^2)}$, where $\mu$ and $\sigma^2$ are computed by a neural component. A unimodal Gaussian distribution encourages each attention head to attend to a single region, as opposed to scattering its attention through many places, and enables tractable computation.
Finally, having $p$, we can compute the context vector $c$ as:
\begin{equation}
\label{eq:context}
c=\mathbb{E}_{p}\left[\bar{X}(t)\right].
\end{equation}

\citet{martins2020sparse} introduced the continuous attention framework for RNNs. In the following section (\S\ref{sec:ltm}), we will explain how it can be used for transformer multi-head attention.

\section{Infinite Memory Transformer}
\begin{figure}[t]
    \centering
    \includegraphics[width=\columnwidth]{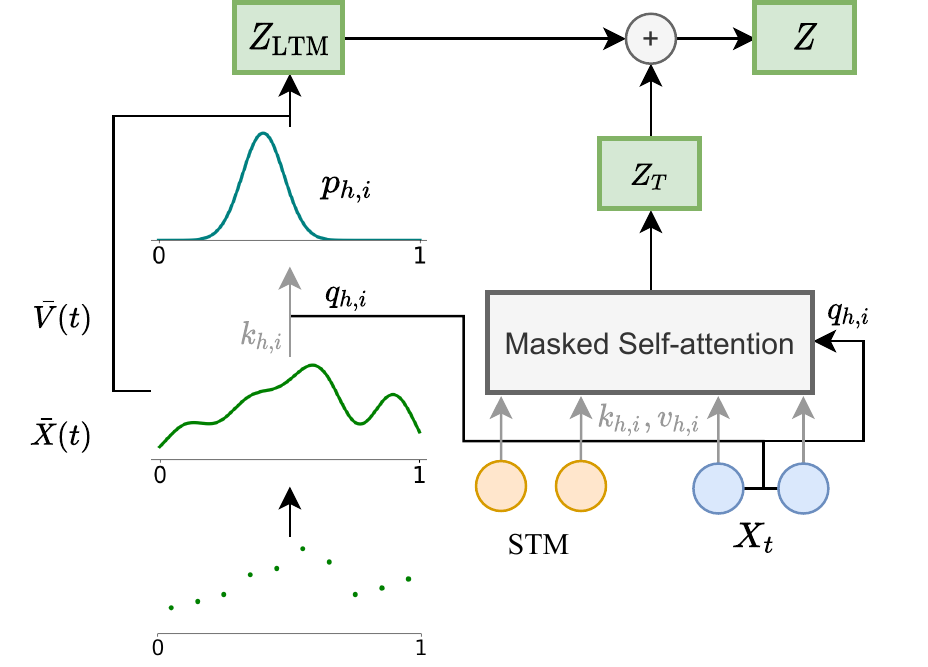}
    \caption{\mbox{$\infty$-former's} attention diagram with sequence of text, $X_t$, of size $L=2$ and STM of size $L_\mathrm{STM}=2$. Circles represent input embeddings or hidden states (depending on the layer) for head $h$ and query $i$. Both the self-attention and the attention over the LTM are performed in parallel for each head and query.}
    \label{fig:infinite_transformer}
\end{figure}

To allow the model to access long-range context, we propose extending the vanilla transformer with a continuous LTM, which stores the input embeddings and hidden states of the previous steps. 
We also consider the possibility of having two memories: the LTM and a short-term memory (STM), which consists in an extension of the transformer's hidden states and is attended to by the transformer's self-attention, as in the transformer-XL \citep{dai2019transformer}.
A diagram of the model is shown in Fig. \ref{fig:infinite_transformer}. 

\subsection{Long-term Memory}
\label{sec:ltm}

For simplicity, let us first assume that the long-term memory contains an explicit input discrete sequence $X$ that consists of the past text sequence's input embeddings or hidden states,\footnote{We stop the gradient with respect to the word embeddings or hidden states before storing them in the LTM.} depending on the layer\footnote{Each layer of the transformer has a different LTM.} (we will later extend this idea to an unbounded memory in \S\ref{sec:unbounded_memory}).

First, we need to transform $X$ into a continuous approximation $\bar{X}(t)$. 
We compute $\bar{X}(t)$ as:
\begin{equation}
    \bar{X}(t) = B^\top \psi(t), 
\end{equation}
where ${\psi(t) \in \mathbb{R}^N}$ are basis functions and coefficients ${B \in \mathbb{R}^{N\times e}}$ are computed as in Eq. \ref{eq:B},
    $B^\top = X^\top G$.
Then, we can compute the LTM keys, ${K_h \in \mathbb{R}^{N \times d}}$, and values, ${V_h \in \mathbb{R}^{N \times d}}$, for each attention head $h$, as: 
\begin{align}
    K_h = B_h W^{K_h},\;\;\; V_h = B_h W^{V_h},
\end{align}
where ${W^{K_h}, W^{V_h} \in \mathbb{R}^{d\times d}}$  are learnable projection matrices.\footnote{Parameter weights are not shared between layers.}
For each query ${q_{h,i} \text{ for } i \in \{1,\dots,L\}}$, we use a parameterized network, which takes as input the attention scores, to compute ${\mu_{h,i} \in ]0,1[}$
and ${\sigma_{h,i}^2 \in \mathbb{R}_{>0}}$:
\begin{align}\label{eq:mu}
    \mu_{h,i} \!&=\! \mathrm{sigmoid}\left(\mathrm{affine}\left( \frac{K_{h}\; q_{h,i}}{\sqrt{d}}\right)\right)\\\label{eq:sigma}
    \sigma^2_{h,i} \!&=\! \mathrm{softplus}\left(\mathrm{affine}\left( \frac{K_{h}\; q_{h,i}}{\sqrt{d}}\right)\right).
\end{align}

Then, using the continuous softmax transformation \citep{martins2020sparse}, we obtain the probability density $p_{h,i}$ as ${\mathcal{N}(t; \mu_{h,i}, \sigma_{h,i}^2)}$.

Finally, having the value function $\bar{V}_{h}(t)$ given as
    ${\bar{V}_{h}(t)= V_{h}^\top  \psi(t),}$
we compute the head-specific representation vectors as in Eq.~\ref{eq:context}:
\begin{equation}
    z_{h,i} = \mathbb{E}_{p_{h,i}}[\bar{V}_{h}] = V_{h}^\top \mathbb{E}_{p_{h,i}}[\psi(t)]
\end{equation}
which form the rows of matrix ${Z_{{\mathrm{LTM}},h} \in \mathbb{R}^{L \times d}}$ that goes through an affine transformation, ${Z_{\mathrm{LTM}} = [Z_{{\mathrm{LTM}},1},\dots,Z_{{\mathrm{LTM}},{H}}]W^O}$.

The long-term representation, $Z_{\mathrm{LTM}}$, is then summed to the transformer context vector, $Z_{\mathrm{T}}$
, to obtain the final context representation ${Z \in \mathbb{R}^{L\times e}}$:
\begin{equation}
    Z =  Z_{\mathrm{T}} + Z_{\mathrm{LTM}},
\end{equation}
which will be the input to the feed-forward layer.

\begin{figure*}[t]
    \centering
    \includegraphics[width=\textwidth]{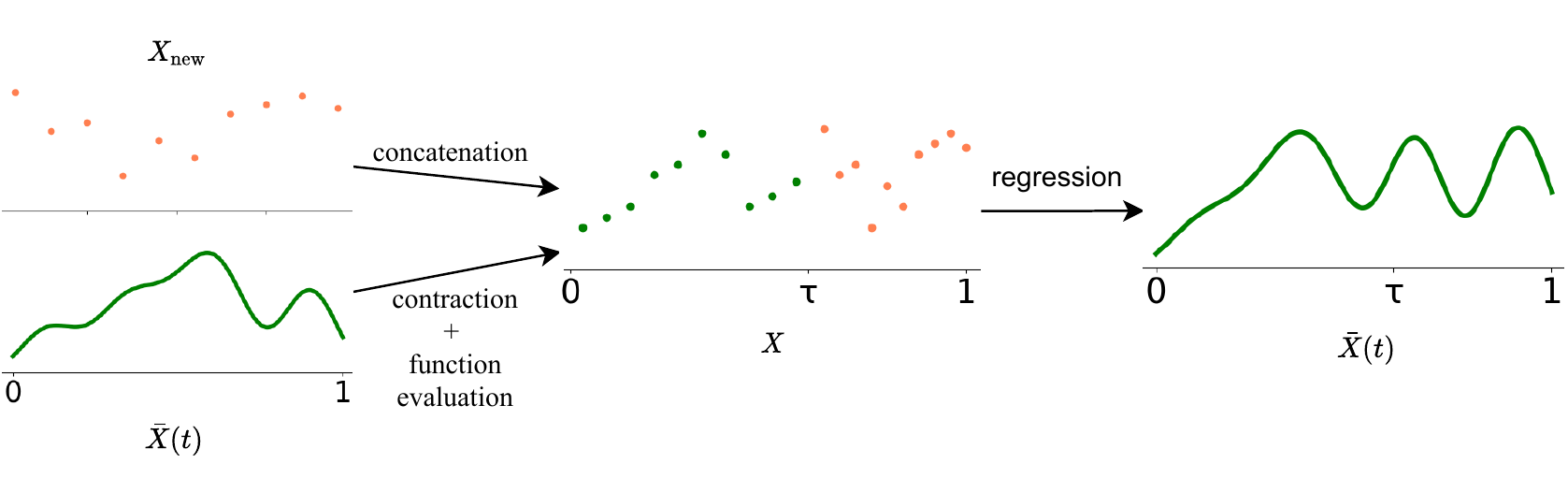}
    \caption{Diagram of the unbounded memory update procedure. This is performed in parallel for each embedding dimension, and repeated throughout the input sequence. We propose two alternatives to select the positions used for the function evaluation: linearly spaced or sticky memories.}
    \label{fig:unbounded_memory}
\end{figure*}

\subsubsection{Attention Complexity}
\label{sec:complexity}
As the $\infty$-former makes use of a continuous-space attention framework \citep{martins2020sparse} to attend over the LTM signal, its key matrix size ${K_{h}\in \mathbb{R}^{N \times d}}$ depends only on the number of basis functions $N$, but \textit{not} on the length of the context being attended to. Thus, the \mbox{$\infty$-former's} attention complexity is also independent of the context's length. 
It corresponds to  ${\mathcal{O}(L\times (L+L_\mathrm{STM}) + L\times N)}$ when also using a short-term memory and ${\mathcal{O}(L^2 + L\times N)}$ when only using the LTM, both $\ll \mathcal{O}(L \times (L+L_{\mathrm{LTM}}))$, which would be the complexity of a vanilla transformer attending to the same context. 
For this reason, the $\infty$-former can attend to arbitrarily long contexts without increasing the amount of computation needed.

\subsection{Unbounded Memory}
\label{sec:unbounded_memory}

When representing the memory as a discrete sequence, to extend it, we need to store the new hidden states in memory. In a vanilla transformer, this is not feasible for long contexts due to the high memory requirements.
However, the $\infty$-former can attend to unbounded context without increasing memory requirements by using continuous attention, as next described and shown in Fig. \ref{fig:unbounded_memory}. 

To be able to build an unbounded representation, we first sample $M$ locations in $[0,1]$ and evaluate $\bar{X}(t)$ at those locations. These locations can simply be linearly spaced, or sampled according to the region importance, as described in \S\ref{sec:sticky_mem}.

Then, we concatenate the corresponding vectors with the new vectors coming from the short-term memory. 
For that, we first need to contract this function by a factor of $\tau \in  \left]0,1\right[$ to make room for the new vectors. We do this by defining: 
\begin{equation}
X^{\mathrm{contracted}}(t) = X(t/\tau) = B^\top \psi(t/\tau).
\end{equation}
Then, we can evaluate $\bar{X}(t)$ at the $M$ locations ${0 \le t_1, t_2, \ldots, t_M \le \tau}$ as:
\begin{equation}\label{eq:eval_signal_LTM}
x_m = B^\top \psi(t_m/\tau), \quad \text{for ${m \in [M]}$},
\end{equation}
and define a matrix  $X_\mathrm{past} = [x_1, x_2, \dots, x_M]^\top \in \mathbb{R}^{M \times e}$ with these vectors as rows. 
After that, we concatenate this matrix with the new vectors $X_\mathrm{new}$, obtaining:
\begin{equation}
    X = \left[{X_\mathrm{past}}^\top, {X_\mathrm{new}}^\top\right]^\top \in \mathbb{R}^{(M+L)\times e}.
\end{equation}

Finally, we simply need to 
perform multivariate ridge regression to compute the new coefficient matrix ${B \in \mathbb{R}^{N \times e}}$, via ${B^\top = X^\top G}$, as in Eq. \ref{eq:B}.
To do this, we need to associate the vectors in $X_\mathrm{past}$ with positions in $[0,\tau]$ and in $X_\mathrm{new}$ with positions in $]\tau,1]$ so that we obtain the matrix ${G \in \mathbb{R}^{(M+L)\times N}}$.
We consider the vectors positions to be linearly spaced.

\subsection{Sticky Memories}
\label{sec:sticky_mem}
When extending the LTM, we evaluate its current signal at $M$ locations in $[0,1]$, as shown in Eq.~\ref{eq:eval_signal_LTM}. These locations can be linearly spaced. However, some regions of the signal can be more relevant than others, and should consequently be given a larger ``memory space'' in the next step LTM's signal. 
To take this into account, we propose sampling the $M$ locations according to the signal's relevance at each region (see Fig.~\ref{fig:sticky_mems_scheme} in App.~\ref{app:sticky_mems}). 
To do so, we construct a histogram based on the attention given to each interval of the signal on the previous step. For that, we first divide the signal into $D$ linearly spaced bins ${\{d_1,\dots,d_D\}}$. Then, we compute the probability given to each bin, $p(d_j)$ for ${j \in \{1,\dots,D\}}$, as:
\begin{equation}
    \label{eq:sm}
    p(d_j) \propto \sum_{h=1}^H \sum_{i=1}^L \int_{d_j} \mathcal{N} (t; \mu_{h,i}, \sigma_{h,i}^2 )\; dt,
\end{equation}
where $H$ is the number of attention heads and $L$ is the sequence length. Note that Eq.~\ref{eq:sm}'s integral can be evaluated efficiently using the erf function:
\begin{equation}
    \int_a^b \mathcal{N}(t; \mu, \sigma^2) = \frac{1}{2} \! \left( \mathrm{erf} \!\left(\dfrac{b}{\sqrt{2}}\right)-\mathrm{erf} \!\left(\dfrac{a}{\sqrt{2}}\right)\right).
\end{equation}
Then, we sample the $M$ locations at which the LTM's signal is evaluated at, according to $p$. 
By doing so, we evaluate the LTM's signal at the regions which were considered more relevant by the previous step's attention, and will, consequently attribute a larger space of the new LTM's signal to the memories stored in those regions.

\subsection{Implementation and Learning Details}

Discrete sequences can be highly irregular and, consequently, difficult to convert into a continuous signal through regression. 
Because of this, before applying multivariate ridge regression to convert the discrete sequence $X$ into a continuous signal,  we use a simple convolutional layer (with $\text{stride}=1$ and $\text{width}=3$)  as a gate, to smooth the sequence:
\begin{equation}
    \Tilde{X} = \mathsf{sigmoid} \left(\mathrm{CNN}(X)\right) \odot X.
\end{equation}

To train the model we use the cross entropy loss. Having a sequence of text $X$ of length $L$ as input, a language model outputs a probability distribution of the next word ${p( x_{t+1}\mid x_{t},\dots,x_{t-L})}$. Given a corpus of $T$ tokens, we train the model to minimize its negative log likelihood:
\begin{equation}
    \mathcal{L}_{\mathrm{NLL}} = -\sum_{t=0}^{T-1} \log p( x_{t+1}\mid x_{t},\dots,x_{t-L}).
\end{equation}

Additionally, in order to avoid having uniform distributions over the LTM, we regularize the continuous attention given to the LTM, by minimizing the Kullback-Leibler (KL) divergence, $D_{\mathrm{KL}}$, between the attention probability density, ${\mathcal{N}(\mu_{h},\sigma_{h})}$, and a Gaussian prior, ${\mathcal{N}(\mu_0,\sigma_0)}$. As different heads can attend to different regions, we set $\mu_0=\mu_h$, regularizing only the attention variance, and get:
\begin{align}
    \!\!\mathcal{L}_\mathrm{KL} \!&=\! \sum_{t=0}^{T-1} \sum_{h=1}^{H} D_\mathrm{KL}\left( \mathcal{N}(\mu_{h},\sigma_{h})\mid \mid \mathcal{N}(\mu_h,\sigma_0) \right)\\
    &=\! \sum_{t=0}^{T-1} \sum_{h=1}^{H} \frac{1}{2} \!\left(\frac{\sigma_h^2}{\sigma_0^2} - \log \! \left(\frac{\sigma_h}{\sigma_0}\right)-1\right).
\end{align}

Thus, the final loss that is minimized corresponds to:
\begin{equation}
    \mathcal{L} = \mathcal{L}_{\mathrm{NLL}} + \lambda_{\mathrm{KL}} \mathcal{L}_\mathrm{KL},
\end{equation}
where $\lambda_{\mathrm{KL}}$ is a hyperparameter that controls the amount of KL regularization.

\section{Experiments}
\label{sec:experiments}
To understand if the $\infty$-former is able to model long contexts, we first performed experiments on a synthetic task, which consists of sorting tokens by their frequencies in a long sequence (\S \ref{sec:sorting}). 
Then, we performed experiments on language modeling  (\S \ref{sec:gpt2}) and document grounded dialogue generation (\S \ref{sec:dgg}) by fine-tuning a pre-trained language model.\footnote{See App.\ref{sec:lm} for further experiments on language modeling.}

\subsection{Sorting}
\label{sec:sorting}

\begin{figure*}[t]
    \centering
    \includegraphics[width=7.9cm]{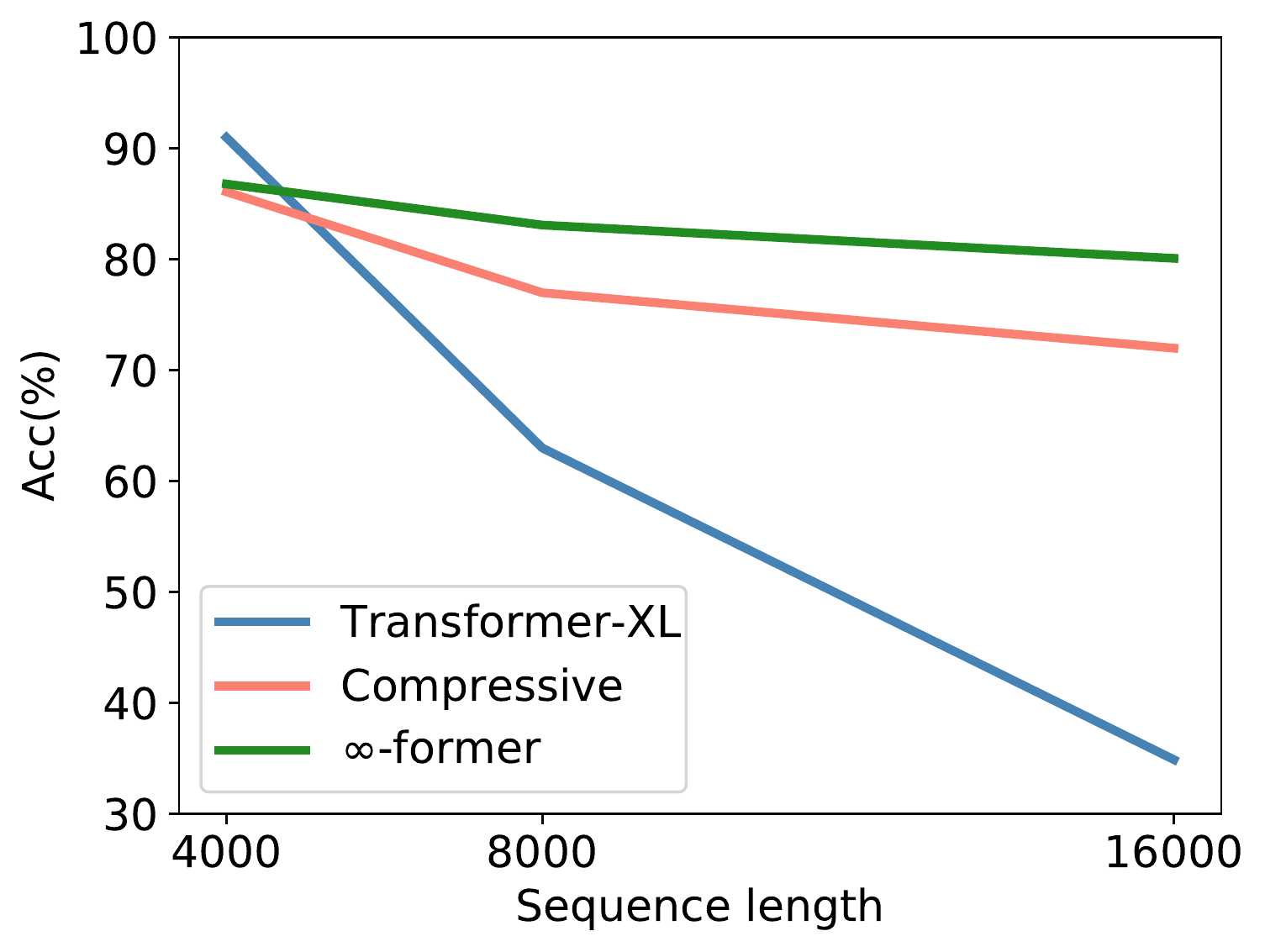}
    \includegraphics[width=7.9cm]{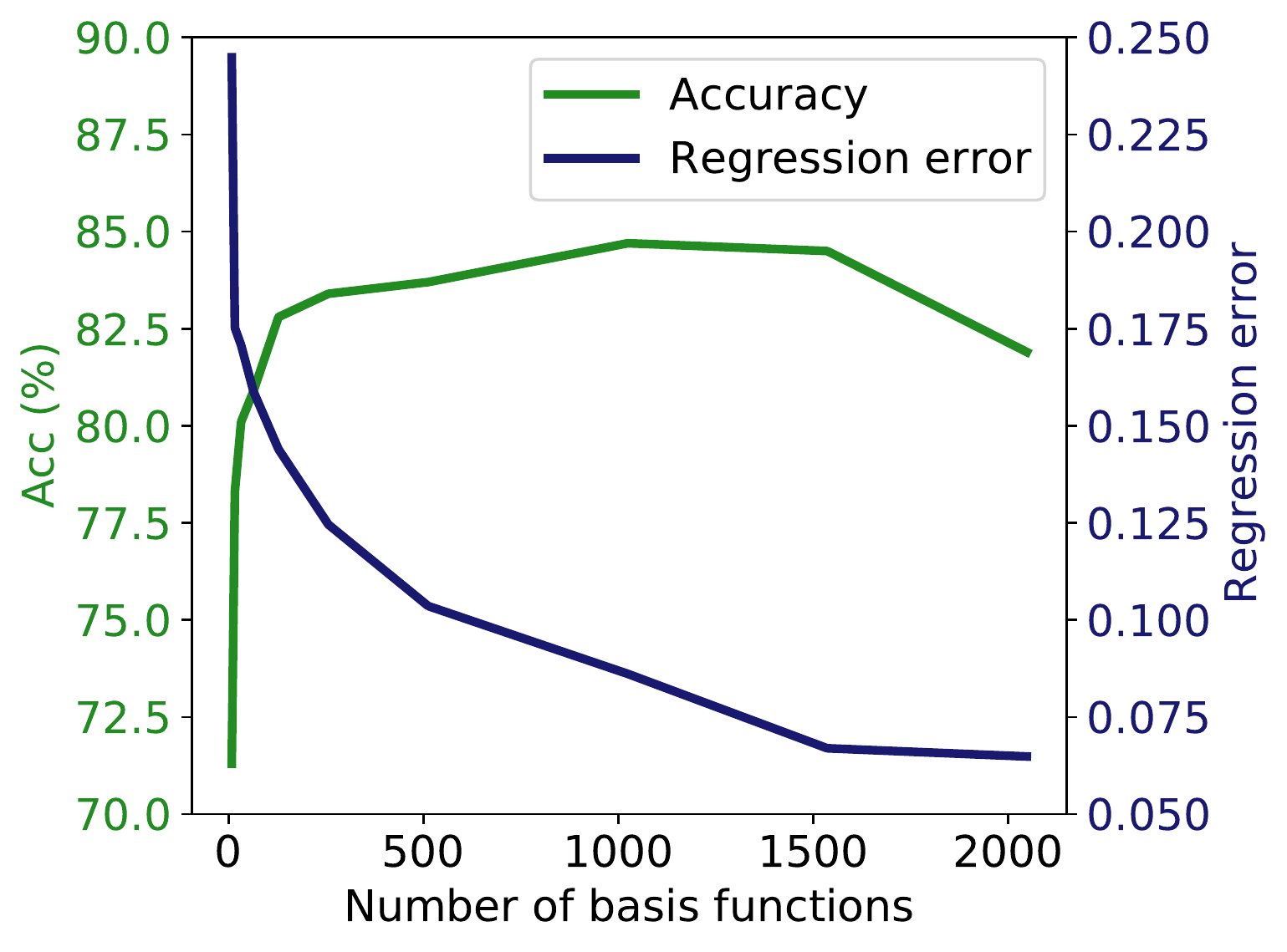}
    \caption{Left: Sorting task accuracy for sequences of length 4,000, 8,000, and 16,000. Right: Sorting task accuracy vs regression mean error, when varying the number of basis functions, for sequences of length 8,000.}
    \label{fig:sorting_results}
\end{figure*}

In this task, the input consists of a sequence of tokens sampled according to a token probability distribution (which is not known to the system). The goal 
is to generate the tokens in the decreasing order of their frequencies in the sequence. 
One example can be:
\begin{equation}
    \underbrace{1\;\; 2\;\; 1\;\; 3\;\; 1\;\; 0\;\; 3\;\; 1\;\; 3\;\; 2\;\;}_{\text{$1$ occurs 4 times; $3$ occurs 3 times, etc.}} \textsc{<sep>}\;\; 1\;\; 3\;\; 2\;\; 0 \nonumber
\end{equation}
To understand if the long-term memory is being effectively used and the transformer is not only performing sorting by modeling the most recent tokens, we design the token probability distribution to \emph{change over time}: namely, we set it as a mixture of two  distributions, ${p = \alpha p_0 + (1-\alpha) p_1}$, where the mixture coefficient $\alpha \in [0,1]$ is progressively increased from 0 to 1 as the sequence is generated. 
The vocabulary has 20 tokens and we experiment with sequences of length 4,000, 8,000, and 16,000. 

\paragraph{Baselines. }
We consider the transformer-XL\footnote{We use the authors' implementation available at \url{https://github.com/kimiyoung/transformer-xl}.} \citep{dai2019transformer} and the compressive transformer\footnote{We use our implementation of the model.} \citep{rae2019compressive} as baselines. The transformer-XL consists of a vanilla transformer \citep{vaswani2017attention} extended with a short-term memory which is composed of the hidden states of the previous steps. The compressive transformer is an extension of the transformer-XL: besides the short-term memory, it has a compressive long-term memory which is composed of  the old vectors of the short-term memory, compressed using a CNN. 
Both the transformer-XL and the compressive transformer require relative positional encodings. 
In contrast, there is no need for positional encodings in the memory in our approach since the memory vectors represent basis coefficients in a predefined continuous space. 

For all models we used a transformer with 3 layers and 6 attention heads, input size $L=1,024$ and memory size 2,048. For the compressive transformer, both memories have size 1,024. For the \mbox{$\infty$-former}, we also consider a STM of size 1,024 and a LTM with $N=1,024$ basis functions, having the models the same computational cost.
Further details are described in App. \ref{app:sorting}.

\paragraph{Results. }
As can be seen in the left plot of Fig.~\ref{fig:sorting_results}, the transformer-XL achieves a slightly higher accuracy than the compressive transformer and  \mbox{$\infty$-former} for a short sequence length (4,000). This is because the transformer-XL is able to keep almost the entire sequence in memory.
However, its accuracy degrades rapidly when the sequence length is increased. 
Both the compressive transformer and \mbox{$\infty$-former} also lead to smaller accuracies when increasing the sequence length, as expected. However, this decrease is not so significant for the \mbox{$\infty$-former}, which indicates that it is better at modeling long sequences.

\paragraph{Regression error analysis. }
To better understand the trade-off between the \mbox{$\infty$-former}'s memory precision and its computational efficiency, we analyze how its regression error and sorting accuracy vary when varying the number of basis functions used, on the sorting task with input sequences of length 8,000. 
As can be seen in the right plot of Fig.~\ref{fig:sorting_results}, the sorting accuracy is negatively correlated with the regression error, which is positively correlated with the number of basis functions.
It can also be observed, that when increasing substantially the number of basis functions the regression error reaches a plateau and the accuracy starts to drop. We posit that the latter is caused by the model having a harder task at selecting the locations it should attend to. 
This shows that, as expected, when increasing \mbox{$\infty$-former}'s efficiency or increasing the size of context being modeled, the memory loses precision.

\subsection{Language Modeling}
\label{sec:gpt2}
To understand if long-term memories can be used to extend a pre-trained language model, we fine-tune GPT-2 small  \citep{radford2019language} on Wikitext-103 \citep{merity2016pointer} and a subset of PG-19 \citep{rae2019compressive} containing the first 2,000 books ($\approx$ 200 million tokens) 
of the training set. To do so, we extend GPT-2 with a continuous long-term memory (\mbox{$\infty$-former}) and a compressed memory (compressive transformer) with a positional bias, based on \citet{press2021train}.\footnote{The compressive transformer requires relative positional encodings. When using only GPT-2's absolute positional encodings the model gives too much attention to  the compressed memory, and, consequently, diverges. Thus, we adapted it by using positional biases on the attention mechanism.} 

For these experiments, we consider transformers with input size $L=512$, 
for the compressive transformer we use a compressed memory of size 512, and for the \mbox{$\infty$-former} we consider a LTM with $N=512$ Gaussian RBFs and a memory threshold of 2,048 tokens, having the same computational budget for the two models. 
Further details and hyperparameters are described in App. \ref{app:gpt2}.

\paragraph{Results. }
\begin{table}[t]
\centering \small
\setlength{\tabcolsep}{3.5ex}
\begin{tabular}{lcc}
\toprule
& Wikitext-103 & PG19 \\
\midrule
GPT-2 & 16.85 & 33.44 \\
Compressive & 16.87 & 33.09 \\
$\infty$-former & 16.64 & 32.61 \\
$\infty$-former (SM) & \textbf{16.61} & \textbf{32.48} \\
\bottomrule
\end{tabular}
\caption{Perplexity on Wikitext-103 and PG19.} 
\label{table:gpt2}
\end{table}

The results reported in Table~\ref{table:gpt2} show that the \mbox{$\infty$-former} leads to perplexity improvements 
on both Wikitext-103 and PG19, while the compressive transformer only has a slight improvement on the latter. 
The improvements obtained by the \mbox{$\infty$-former} are larger on the PG19 dataset, which can be justified by the nature of the datasets: books have more long range dependencies than Wikipedia articles \citep{rae2019compressive}.

\subsection{Document Grounded Dialogue}
\label{sec:dgg}
In document grounded dialogue generation, besides the dialogue history, models have access to a document concerning the conversation's topic. In the CMU Document Grounded Conversation dataset (CMU-DoG) \citep{zhou2018dataset}, the dialogues are about movies and a summary of the movie is given as the auxiliary document; the auxiliary document is divided into parts that should be considered for the different utterances of the dialogue. In this paper, to evaluate the usefulness of the long-term memories, we make this task slightly more challenging by only giving the models access to the document before the start of the dialogue.

We fine-tune \mbox{GPT-2} small \citep{radford2019language} using an approach based on \citet{wolf2019transfertransfo}. To allow the model to keep the whole document on memory, we extend GPT-2 with a continuous LTM ($\infty$-former) with $N=512$ basis functions.
As baselines, we use GPT-2, with and without access (GPT-2 w/o doc) to the auxiliary document, with  input size $L=512$, and GPT-2 with a compressed memory with attention positional biases (compressive), of size 512. Further details and hyper-parameters are stated in App. \ref{app:dgd}. 

To evaluate the models we use the metrics: perplexity, F1 score, Rouge-1 and Rouge-L \citep{lin2004rouge}, and Meteor \citep{banerjee2005meteor}.

\paragraph{Results. }
\begin{table}[t]
\centering \small
\setlength{\tabcolsep}{0.7ex}
\begin{tabular}{l@{\hspace{1ex}}c@{\hspace{2ex}}cccc}
\toprule
& PPL & F1 & Rouge-1 & Rouge-L & Meteor \\
\midrule
GPT-2 w/o doc & 19.43 & 7.82 & 12.18 & 10.17 & 6.10    \\
GPT-2 & 18.53 & 8.64 & 14.61 & 12.03 & 7.15  \\
Compressive & \textbf{18.02} & 8.78 & 14.74 & 12.14 & 7.29  \\
$\infty$-former & \textbf{18.02} & 8.92 & 15.28 & 12.51 & 7.52  \\
$\infty$-former (SM) & 18.04 & \textbf{9.01} & \textbf{15.37} & \textbf{12.56} & \textbf{7.55} \\
\bottomrule
\end{tabular}
\caption{Results on CMU-DoG dataset.}
\label{table:dgg}
\end{table}

As shown in Table \ref{table:dgg}, by keeping the whole auxiliary document in memory, the \mbox{$\infty$-former} and the compressive transformer are able to generate better utterances, according to all metrics. 
While the compressive and \mbox{$\infty$-former} achieve essentially the same perplexity in this task, 
the \mbox{$\infty$-former} achieves consistently better scores on all other metrics. 
Also, using sticky memories leads to slightly better results on those metrics, which suggests that attributing a larger space in the LTM to the most relevant tokens can be beneficial.

\paragraph{Analysis. } 
\begin{figure*}[t]
  \centering
    \includegraphics[width=\textwidth]{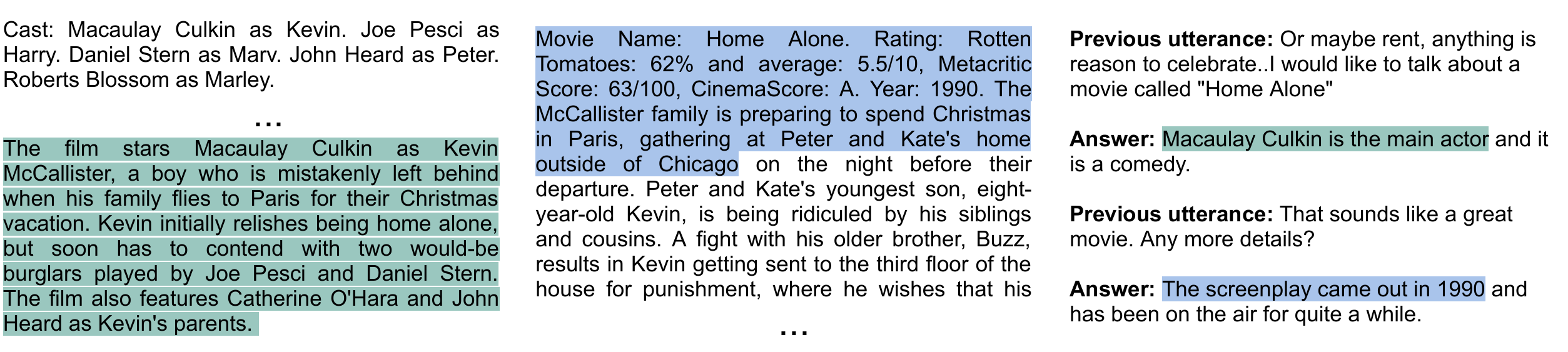}
  \caption{Examples of answers generated by \mbox{$\infty$-former} on a dialogue about the movie ``Home Alone''. The excerpts from the LTM that are more attended to throughout the utterances generation are highlighted on each color, correspondingly.}
  \label{fig:example_dgg}
\end{figure*}
\begin{figure*}[t]
  \centering
    \includegraphics[width=\textwidth]{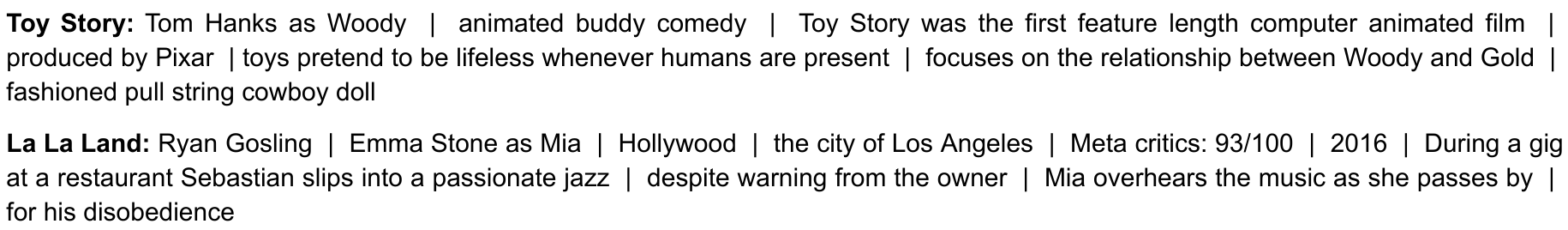}
  \caption{Phrases that hold larger spaces of the LTM, when using sticky memories, for two dialogue examples (in App. \ref{sec:examples}).}
  \label{fig:example_dgg_sm}
\end{figure*}
In Fig. \ref{fig:example_dgg}, we show examples of utterances generated by \mbox{$\infty$-former} along with the excerpts from the LTM that receive higher attention throughout the utterances' generation. In these examples, we can clearly see that these excerpts are highly pertinent to the answers being generated. Also, in Fig. \ref{fig:example_dgg_sm}, we can see that the phrases which are attributed larger spaces in the LTM, when using sticky memories, are relevant to the conversations.

\section{Related Work}
\paragraph{Continuous attention. }
\citet{martins2020sparse} introduced 1D and 2D continuous attention, using Gaussians and truncated parabolas as densities. They applied it to RNN-based document classification, machine translation, and visual question answering.
Several other works have also proposed the use of (discretized) Gaussian attention for natural language processing tasks: \citet{guo2019gaussian} proposed a Gaussian prior to the self-attention mechanism to bias the model to give higher attention to nearby words, and applied it to natural language inference;
\citet{you2020hard} proposed the use of hard-coded Gaussian attention as input-agnostic self-attention layer for machine translation; \citet{dubois2020location} proposed using Gaussian attention as a location attention mechanism to improve the model generalization to longer sequences. However, these approaches still consider discrete sequences and compute the attention by evaluating the Gaussian density at the token positions. 
\citet{Farinhas2021MultimodalCV} extend continuous attention to multimodal densities, \textit{i.e.}, mixtures of Gaussians, and applied it to VQA. In this paper, we opt for the simpler case, an unimodal Gaussian, and leave sparse and multimodal continuous attention for future work.

\paragraph{Efficient transformers. }
Several methods have been proposed that reduce the transformer's attention complexity, and can, consequently, model longer contexts. Some of these do so by performing sparse attention, either by selecting pre-defined attention patterns \citep{child2019generating,beltagy2020longformer,zaheer2020big}, or by learning these patterns from data \citep{kitaev2020reformer,vyas2020fast,tay2020sparse,roy2021efficient,wang2021cluster}. 
Other works focus on directly reducing the attention complexity by 
applying the (reversed) kernel trick 
\citep{katharopoulos2020transformers,choromanski2020rethinking,peng2021random,jaegle2021perceiver}. 
Closer to our approach are the transformer-XL and compressive transformer models \citep{dai2019transformer, rae2019compressive}, which extend the vanilla transformer with a bounded memory.

\paragraph{Memory-augmented language models. }
RNNs, LSTMs, and GRUs \citep{hochreiter1997long,cho2014properties} have the ability of keeping a memory state of the past.  However, these require backpropagation through time which is impractical for long sequences.
Because of this, \citet{graves2014neural}, \citet{weston2014memory}, \citet{joulin2015inferring} and \citet{grefenstette2015learning} proposed extending RNNs with an external memory, while \citet{chandar2016hierarchical} and \citet{rae2016scaling} proposed efficient procedures to read and write from these memories, using hierarchies and sparsity.
\citet{grave2016improving} and \citet{merity2016pointer} proposed the use of cache-based memories which store pairs of hidden states and output tokens from previous steps. The distribution over the words in the memory is then combined with the distribution given by the language model.
More recently, \citet{khandelwal2019generalization} and \citet{yogatama2021adaptive} proposed using nearest neighbors to retrieve words from a key-based memory constructed based on the training data. Similarly, \citet{fan2021augmenting} proposed retrieving sentences from a memory based on the training data and auxiliary information. \citet{khandelwal2019generalization} proposed interpolating the retrieved words probability distributions with the probability over the vocabulary words when generating a new word, while \citet{yogatama2021adaptive} and \citet{fan2021augmenting} proposed combining the information at the architecture level. 
These models have the disadvantage of needing to perform a retrieval step when generating each token / utterance, which can be computationally expensive. 
These approaches are orthogonal to the \mbox{$\infty$-former}'s LTM and in future work the two can be combined.

\section{Conclusions}
In this paper, we proposed the $\infty$-former: a transformer extended with an unbounded long-term memory. By using a continuous-space attention framework, its attention complexity is independent of the context's length, 
which allows the model to attend to arbitrarily long contexts while keeping a fixed computation budget. 
By updating the memory taking into account past usage, the model keeps ``sticky memories'', enforcing the persistence of relevant information in memory over time. 
Experiments on a sorting synthetic task show that $\infty$-former scales up to long sequences, maintaining a high accuracy. Experiments on language modeling and document grounded dialogue generation by fine-tuning a pre-trained language model have shown improvements across several metrics.

\section*{Ethics Statement}
Transformer models that attend to long contexts, to improve their generation quality, need large amounts of computation and memory to perform self-attention. In this paper, we propose an extension to a transformer model that makes the attention complexity independent of the length of the context being attended to. This can lead to a reduced number of parameters needed to model the same context, which can, consequently, lead to gains in efficiency and reduce energy consumption. 

On the other hand, the $\infty$-former, as well as the other transformer language models, can be used on questionable scenarios, such as the generation of fake news \citep{zellers2019defending}, defamatory text \citep{wallace2019universal}, or other undesired content.

\section*{Acknowledgments}
This work was supported by the European Research Council (ERC StG DeepSPIN 758969), 
by the P2020 project MAIA (contract 045909), by the Funda\c{c}\~ao para a Ci\^encia e Tecnologia through project PTDC/CCI-INF/4703/2021 (PRELUNA, contract UIDB/50008/2020), by the EU H2020 SELMA project (grant agreement No 957017), and by contract PD/BD/150633/2020 in the scope of the  Doctoral Program  FCT - PD/00140/2013 NETSyS. We thank Jack Rae, Tom Schaul, the SARDINE team members, and the reviewers for helpful discussion and feedback.

\bibliography{anthology,custom}
\bibliographystyle{acl_natbib}

\clearpage
\appendix

\section{Multivariate ridge regression}
\label{app:reg}
The coefficient matrix $B \in \mathbb{R}^{N\times e}$ is obtained with multivariate ridge regression criterion so that $\bar{X}(t_i) \approx x_i$ for each $i \in [L]$, which leads to the closed form:
\begin{align}\label{eq:reg}
    B^\top &= \argmin_{B^\top} ||B^\top F - X^\top||_\mathcal{F}^2 + \lambda||B||_\mathcal{F}^2 \\\nonumber
        &= X^\top F^\top (FF^\top + \lambda I)^{-1} = X^\top G,
\end{align}
where $F=[\psi(t_1),\dots,\psi(t_L)]$ packs the basis vectors for $L$ locations and $||\cdot||_\mathcal{F}$ is the Frobenius norm. As $G \in \mathbb{R}^{L\times N}$ only depends of $F$, it can be computed offline.

\section{Sticky memories}
\begin{figure*}[t]
  \centering
    \includegraphics[width=12cm]{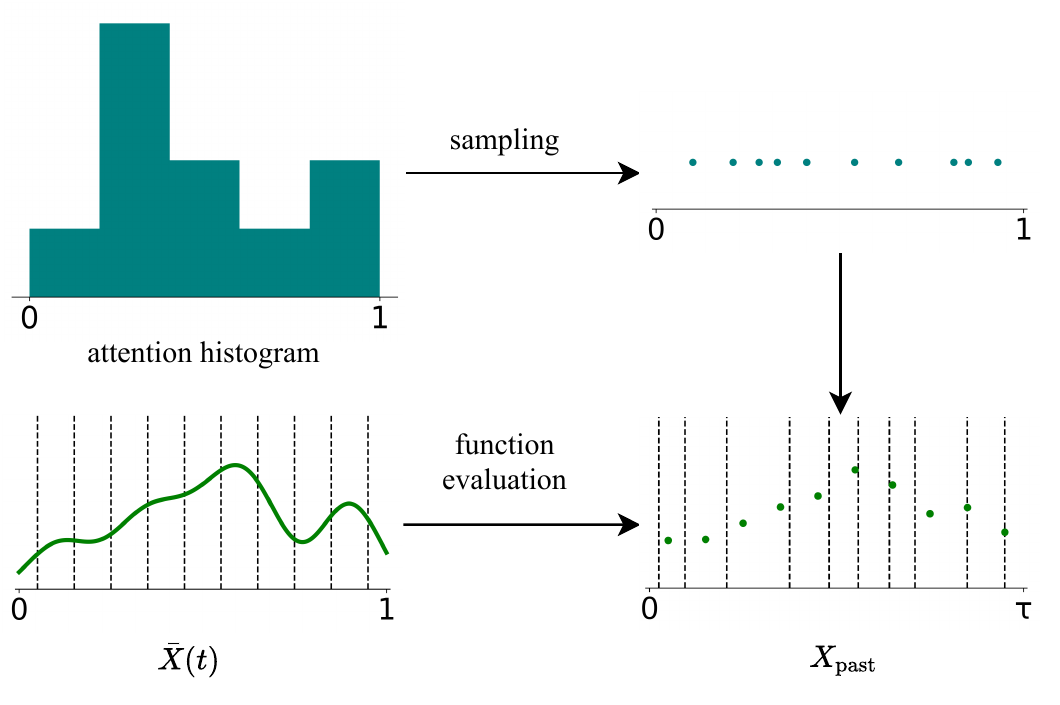}
  \caption{Sticky memories procedure diagram. The dashed vertical lines correspond to the position of the words in the LTM signal.}
  \label{fig:sticky_mems_scheme}
\end{figure*}

We present in Fig. \ref{fig:sticky_mems_scheme} a scheme of the sticky memories procedure.
First we sample $M$ locations from the previous step LTM attention histogram (Eq. \ref{eq:sm}). Then, we evaluate the LTM's signal at the sampled locations (Eq. \ref{eq:eval_signal_LTM}). Finally, we consider that the sampled vectors, $X_\mathrm{past}$, are linearly spaced in $[0,\tau]$. This way, the model is able to attribute larger spaces of its memory to the relevant words.
\label{app:sticky_mems}

\section{Experimental details}
\subsection{Sorting}
\label{app:sorting}
For the compressive transformer, we consider compression rates of size 2 for sequences of length 4,000, from $2$ to $6$ for sequences of length 8,000, and from $2$ to $12$ for sequences of length 16,000. We also experiment training the compressive transformer with and without the attention reconstruction auxiliary loss. For the $\infty$-former we consider 1,024 Gaussian RBFs $\mathcal{N} (t;\Tilde{\mu},\Tilde{\sigma}^2)$ with $\Tilde{\mu}$ linearly spaced in $[0,1]$ and $\Tilde{\sigma} \in \{.01,.05\}$. We set $\tau=0.75$ and for the KL regularization we used $\lambda_{\mathrm{KL}} = 1 \times 10^{-5}$ and $\sigma_0 = 0.05$.

For this task, for each sequence length, we created a training set with 8,000 sequences and validation and test sets with 800 sequences. We trained all models with batches of size 8 for 20 epochs on 1 Nvidia GeForce RTX 2080 Ti or 1 Nvidia GeForce GTX 1080 Ti GPU with $\approx 11$ Gb of memory, using the Adam optimizer \citep{kingma2015adam}. For the sequences of length 4,000 and 8,000 we used a learning rate of $2.5 \times 10^{-4}$ while for sequences of length 16,000 we used a learning rate of $2 \times 10^{-4}$. The learning rate was decayed to 0 until the end of training with a cosine schedule.

\subsection{Pre-trained Language Models}
\label{app:gpt2}
In these experiments, we fine-tune the GPT-2 small, which is composed of 12 layers with 12 attention heads, on the English dataset Wikitext-103 and on a subset of the English dataset PG19\footnote{Dataset available at \href{https://github.com/deepmind/pg19}{https://github.com/deepmind/pg19}.} containing the first 2,000 books. We consider an input size  $L=512$ and a long-term memory with $N=512$ Gaussian RBFs $\mathcal{N} (t;\Tilde{\mu},\Tilde{\sigma}^2)$ with $\Tilde{\mu}$ linearly spaced in $[0,1]$ and $\Tilde{\sigma} \in \{.005,.01\}$ and for the KL regularization we use $\lambda_{\mathrm{KL}} = 1 \times 10^{-6}$ and $\sigma_0 = 0.05$. We set $\tau=0.5$. For the compressive transformer we also consider a compressed memory of size 512 with a compression rate of 4, and  train the model with the auxiliary reconstruction loss.

We fine-tuned GPT-2 small with a batch size of 1 on 1 Nvidia GeForce RTX 2080 Ti or 1 Nvidia GeForce GTX 1080 Ti GPU with $\approx 11$ Gb of memory, using the Adam optimizer \citep{kingma2015adam} for 1 epoch with a learning rate of $5\times 10^{-5}$ for the GPT-2 parameters and a learning rate of $2.5\times 10^{-4}$ for the LTM parameters.

\subsection{Document Grounded Generation}
\label{app:dgd}
In these experiments, we fine-tune the GPT-2 small, which is composed of 12 layers with 12 attention heads, on the English dataset CMU - Document Grounded Conversations\footnote{Dataset available at \href{https://github.com/festvox/datasets-CMU_DoG}{https://github.com/festvox/datasets-CMU\_DoG}.} (CMU-DoG. CMU-DoG has 4112 conversations, being the proportion of train/validation/test split 0.8/0.05/0.15.

We consider an input size $L=512$ and a long-term memory with $N=512$ Gaussian RBFs $\mathcal{N} (t;\Tilde{\mu},\Tilde{\sigma}^2)$ with $\Tilde{\mu}$ linearly spaced in $[0,1]$ and $\Tilde{\sigma} \in \{.005,.01\}$ and for the KL regularization we use $\lambda_{\mathrm{KL}} = 1 \times 10^{-6}$ and $\sigma_0 = 0.05$. We set $\tau=0.5$. For the compressive transformer we consider a compressed memory of size 512 with a compression rate of 3, and  train the model with the auxiliary reconstruction loss.
We fine-tuned GPT-2 small with a batch size of 1 on 1 Nvidia GeForce RTX 2080 Ti or 1 Nvidia GeForce GTX 1080 Ti GPU with $\approx 11$ Gb of memory, using the Adam optimizer \citep{kingma2015adam} with a linearly decayed learning rate of $5\times 10^{-5}$, for 5 epochs.

\section{Additional experiments}
\label{sec:lm}
We also perform language modeling experiments on the Wikitext-103 dataset\footnote{Dataset available at \href{https://blog.einstein.ai/the-wikitext-long-term-dependency-language-modeling-dataset/}{https://blog.einstein.ai/the-wikitext-long-term-dependency-language-modeling-dataset/}.} \citep{merity2016pointer} which has a training set with 103 million tokens and validation and test sets with  217,646 and 245,569 tokens, respectively. 
For that, we follow the standard architecture of the transformer-XL \citep{dai2019transformer}, which consists of a transformer with 16 layers and 10 attention heads. For the transformer-XL, we experiment with a memory of size 150. For the compressive transformer, we consider that both memories have a size of 150 and a compression rate of 4. 
For the \mbox{$\infty$-former} we consider a short-term memory of size 150, a continuous long-term memory with 150 Gaussian RBFs, and a memory threshold of 900 tokens. 

For this experiment, we use a transformer with 16 layers, 10 heads, embeddings of size 410, and  a feed-forward hidden size of 2100. For the compressive transformer, we follow \citet{rae2019compressive} and use a compression rate of 4 and the attention reconstruction auxiliary loss.  For the $\infty$-former we consider 150 Gaussian RBFs $\mathcal{N} (t;\Tilde{\mu},\Tilde{\sigma}^2)$ with $\Tilde{\mu}$ linearly spaced in $[0,1]$ and $\Tilde{\sigma} \in \{.01,.05\}$. 
We set $\tau=0.5$ and for the KL regularization we used $\lambda_{\mathrm{KL}} = 1 \times 10^{-5}$ and $\sigma_0 = 0.1$.

We trained all models, from scratch, with batches of size 40 for 250,000 steps on 1 Nvidia Titan RTX or 1 Nvidia Quadro RTX 6000 with $\approx 24$ GPU Gb of memory using the Adam optimizer \citep{kingma2015adam} with a learning rate of $2.5 \times 10^{-4}$. The learning rate was decayed to 0 until the end of training with a cosine schedule.

\paragraph{Results. }
As can be seen in Table~\ref{table:wt-103}, extending the model with a long-term memory leads to a better perplexity, for both the compressive transformer and  $\infty$-former. 
Moreover, the $\infty$-former slightly outperforms the compressive transformer. We can also see that using sticky memories leads to a somewhat lower perplexity, which shows that it helps the model to focus on the relevant memories.

\begin{table}[t]
\vspace{\baselineskip}
\centering \small
\setlength{\tabcolsep}{1ex}
\begin{tabular}{lccc}
\toprule
& STM & LTM & Perplexity \\
\midrule
Transformer-XL & 150 & ---- & 24.52 \\
Compressive & 150 & 150 & 24.41 \\
$\infty$-former & 150 & 150 & 24.29 \\
$\infty$-former (Sticky memories) & 150 & 150 & \textbf{24.22} \\
\bottomrule
\end{tabular}
\caption{Perplexity on Wikitext-103. }
\label{table:wt-103}
\end{table}

\paragraph{Analysis. }
\label{sec:analysis_lm}

\begin{figure*}[h]
  \centering
    \includegraphics[width=5.2cm]{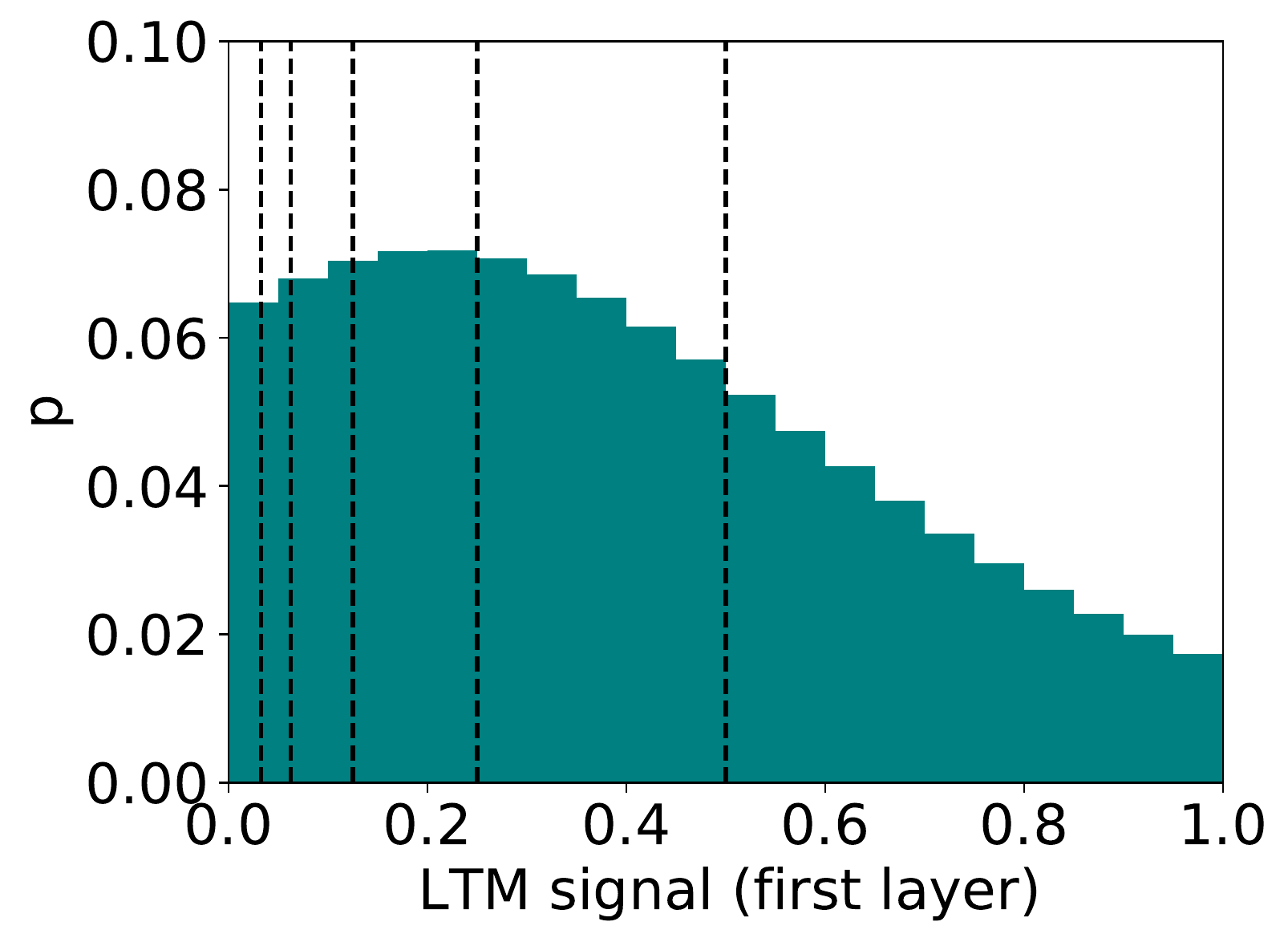}
    \includegraphics[width=5.2cm]{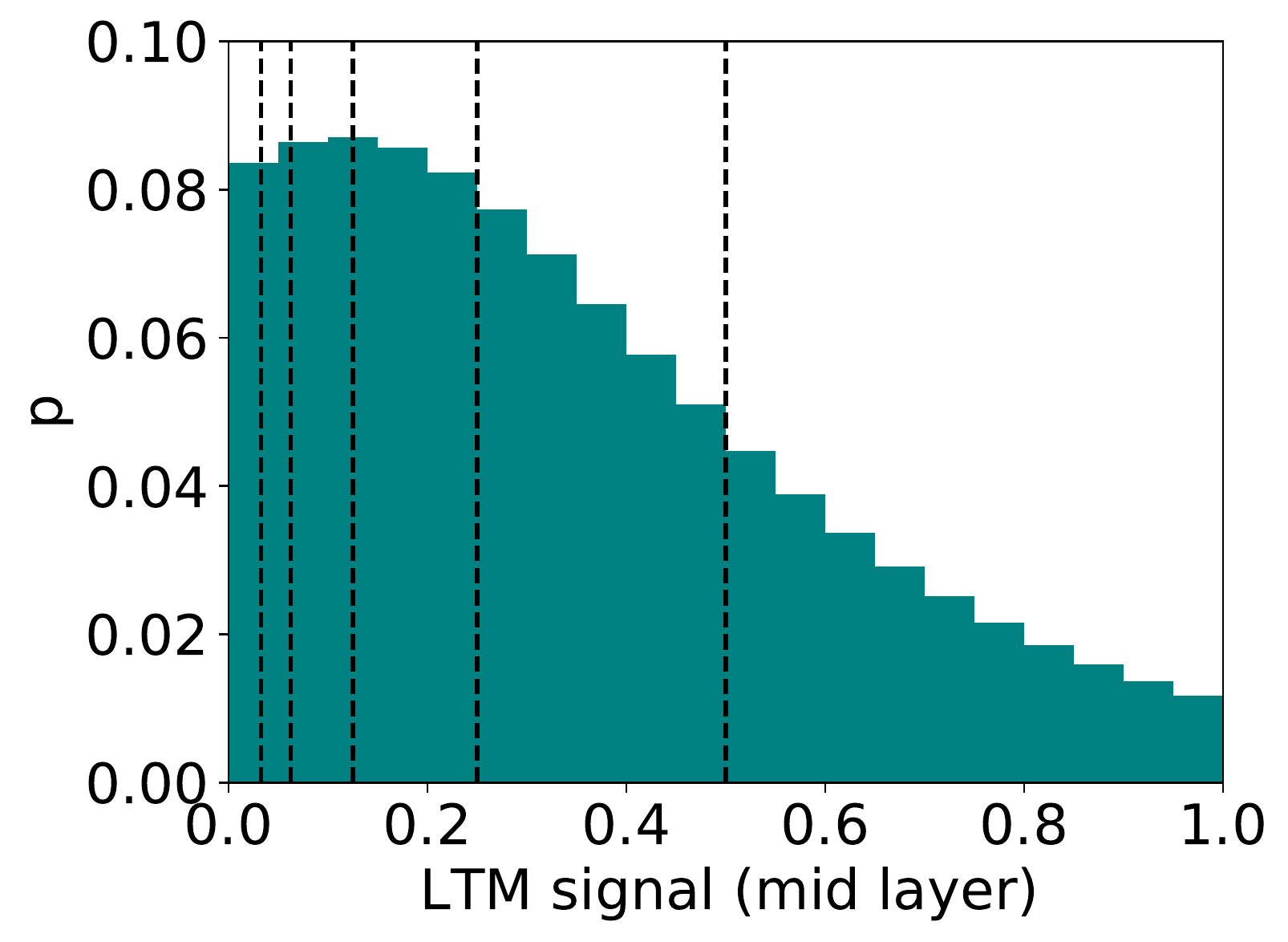}
    \includegraphics[width=5.2cm]{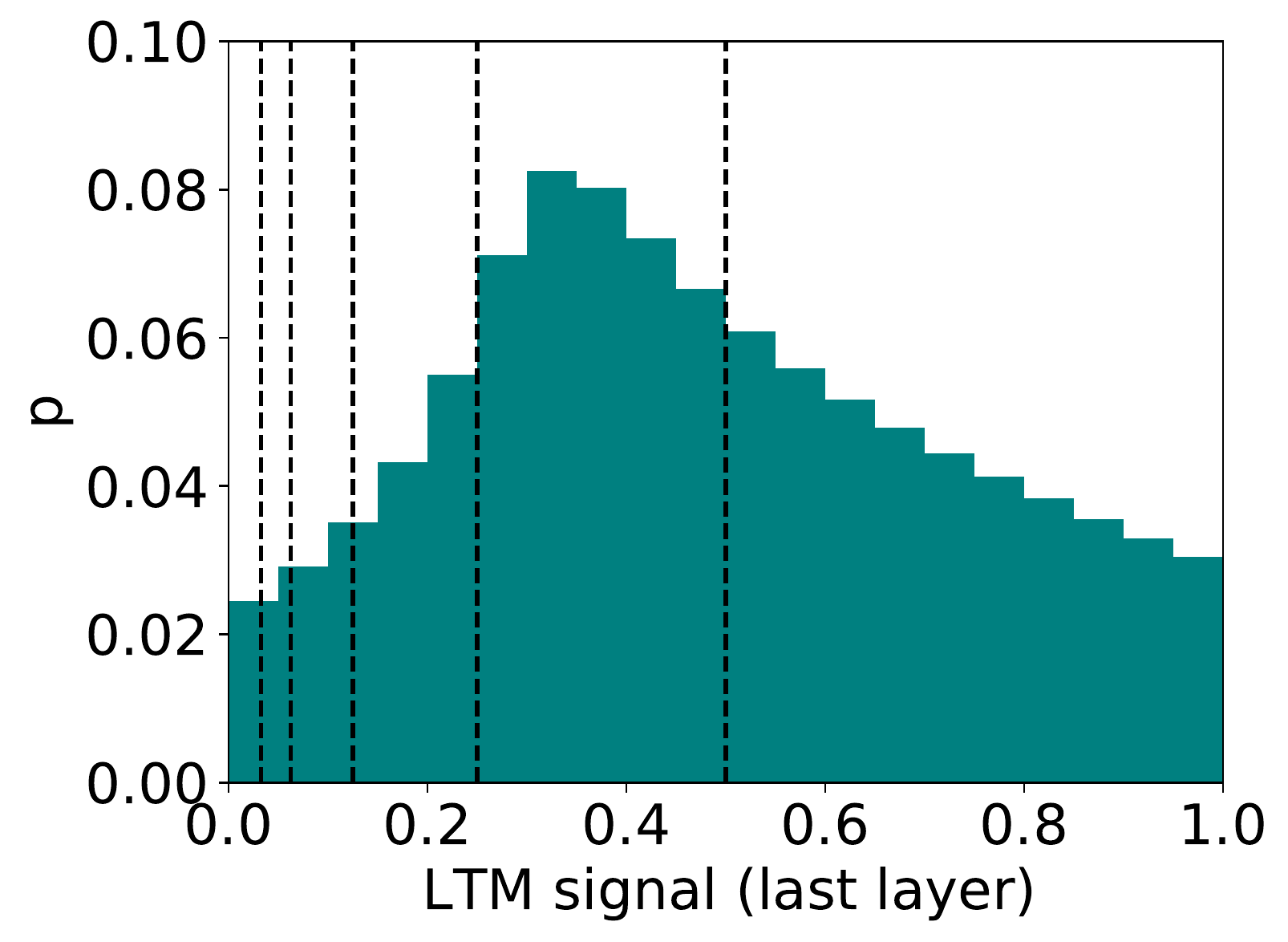}
  \caption{Histograms of attention given to the LTM by $\infty$-former, for the first (on the left), middle (on the middle), and last (on the right) layers. The dashed vertical lines represent the limits of the memory segments ($\tau$) for the various memory updates.}
  \label{fig:hist_attns_inf}
\end{figure*}

\begin{figure*}[h]
  \centering
    \includegraphics[width=\textwidth]{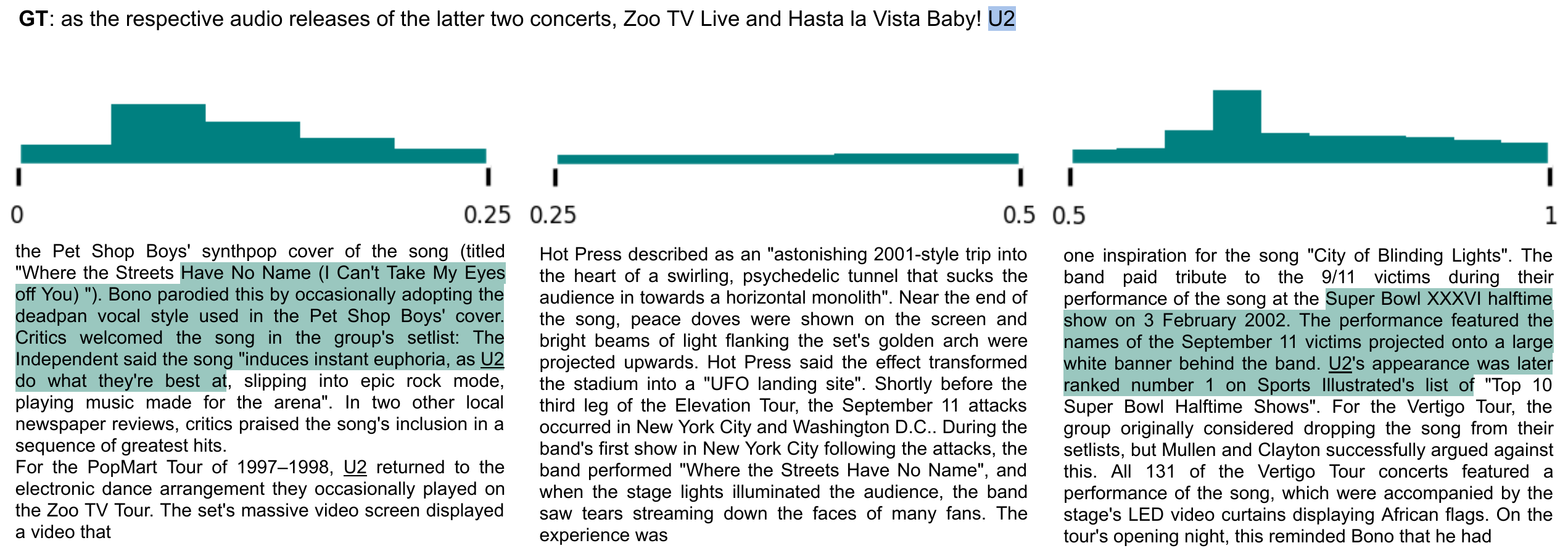}
  \caption{Example of attention given by $\infty$-former to the last layer's long-term memory, when predicting the ground truth word ``U2''. The words in the LTM which receive higher attention ($>0.05$) are shaded.}
  \label{fig:example}
\end{figure*}

\begin{figure*}[h]
  \centering
    \includegraphics[width=\textwidth]{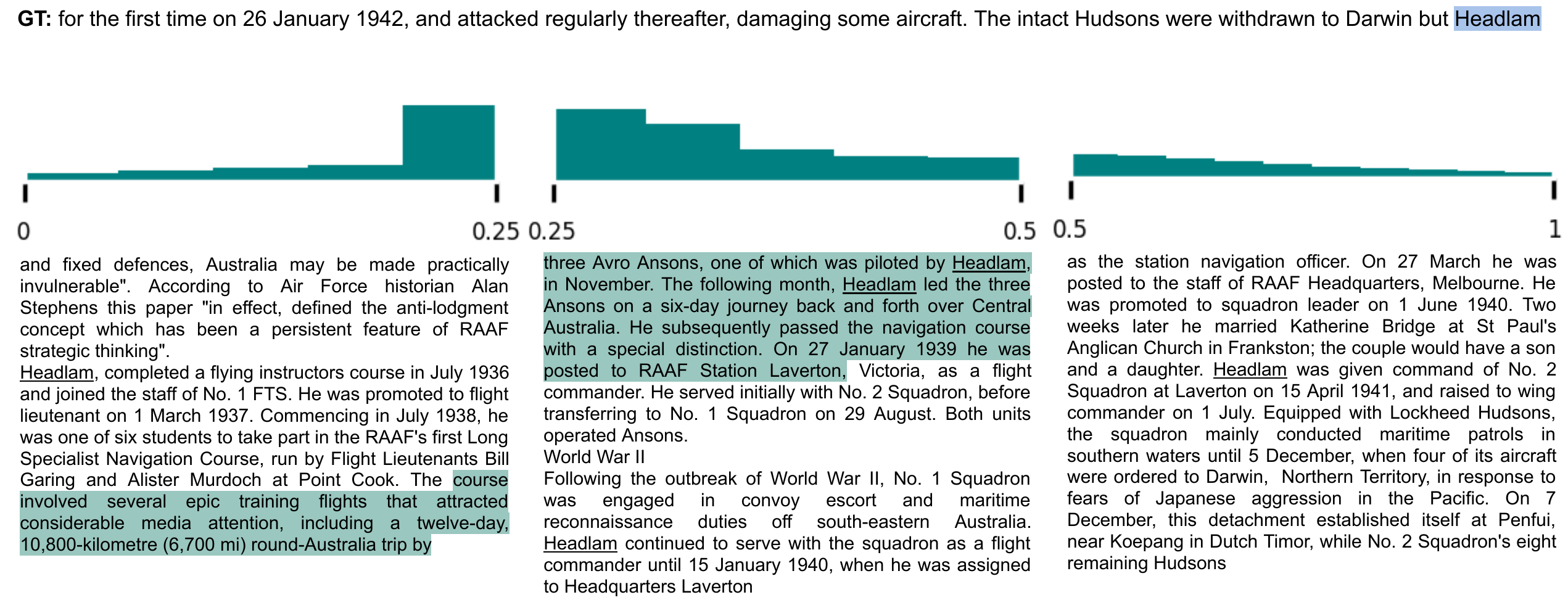}
  \caption{Example of attention given by $\infty$-former to the last layer's long-term memory, when predicting the ground truth word ``Headlam''. The words in the long-term memory which receive higher attention (bigger than $0.05$) are shaded.}
  \label{fig:example_2}
\end{figure*}

To better understand whether $\infty$-former is paying more attention to the older memories in the LTM or to the most recent ones, we plotted a histogram of the attention given to each region of the long-term memory when predicting the tokens on the validation set. As can be seen in Fig.~\ref{fig:hist_attns_inf}, in the first  and middle layers, the $\infty$-former tends to focus more on the older memories, while in the last layer, the attention pattern is more uniform. 
In Figs. \ref{fig:example} and \ref{fig:example_2}, we present examples of words for which the $\infty$-former has lower perplexity than the transformer-XL along with the attention given by the $\infty$-former to the last layer's LTM. We can see that the word being predicted is present several times in the long-term memory and $\infty$-former gives higher attention to those regions.

\begin{figure*}[h]
  \centering
    \includegraphics[width=15cm]{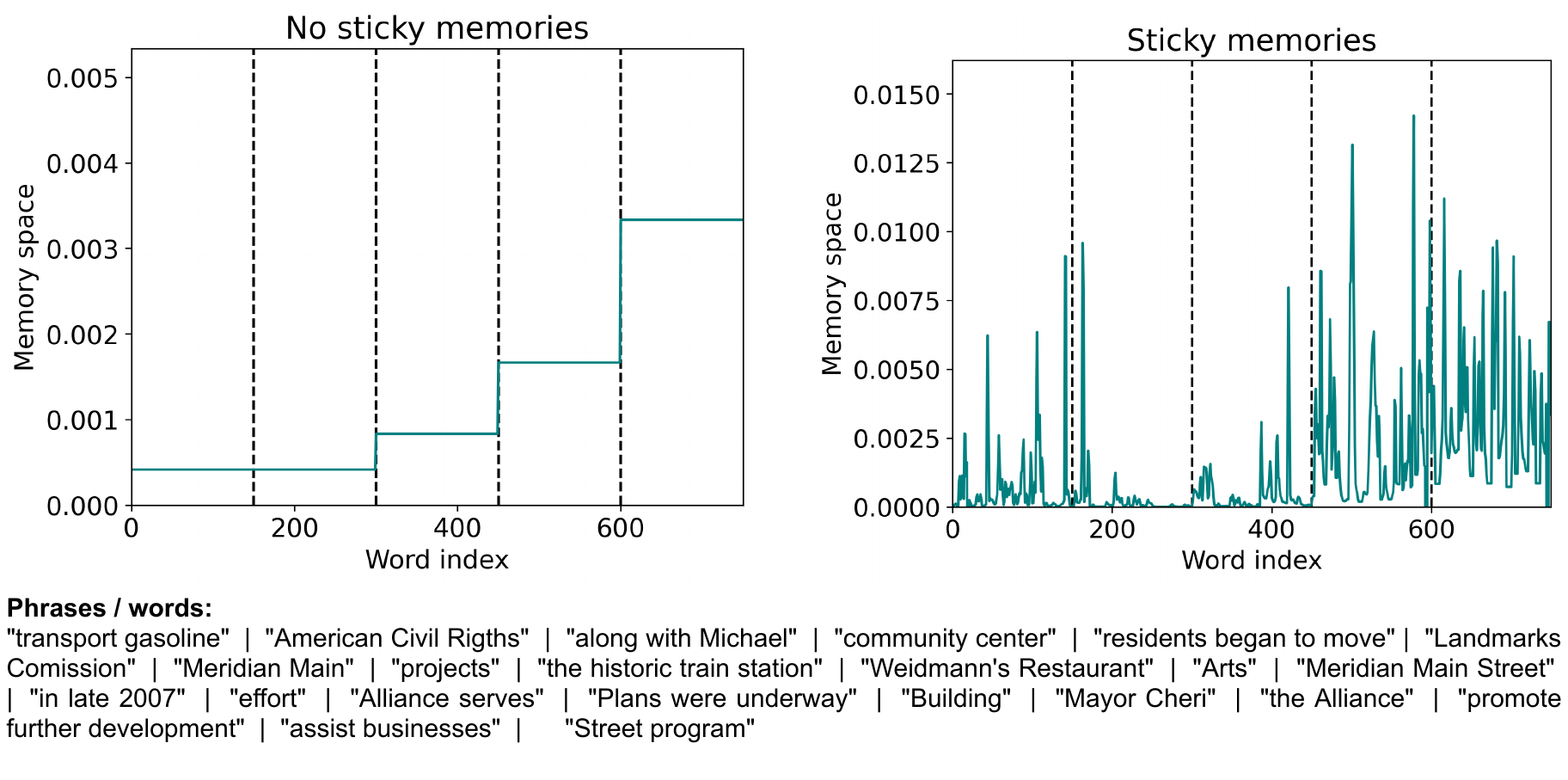}
  \caption{Example of the memory space attributed to each word in the last layer's long-term memory (after 5 updates) without / with the sticky memories procedure, along with the words / phrases which have the largest memory spaces when using sticky memories (top peaks with space$>.005$). Excerpt of the sequence being generated in this example: \textit{``Given Meridian's site as a railroad junction, its travelers have attracted the development of many hotels.''} The dashed vertical lines represent the limits of the memory segments for the various memory updates.}
  \label{fig:example_sticky_mems}
\end{figure*}

\begin{figure*}[t]
  \centering
    \includegraphics[width=15.5cm]{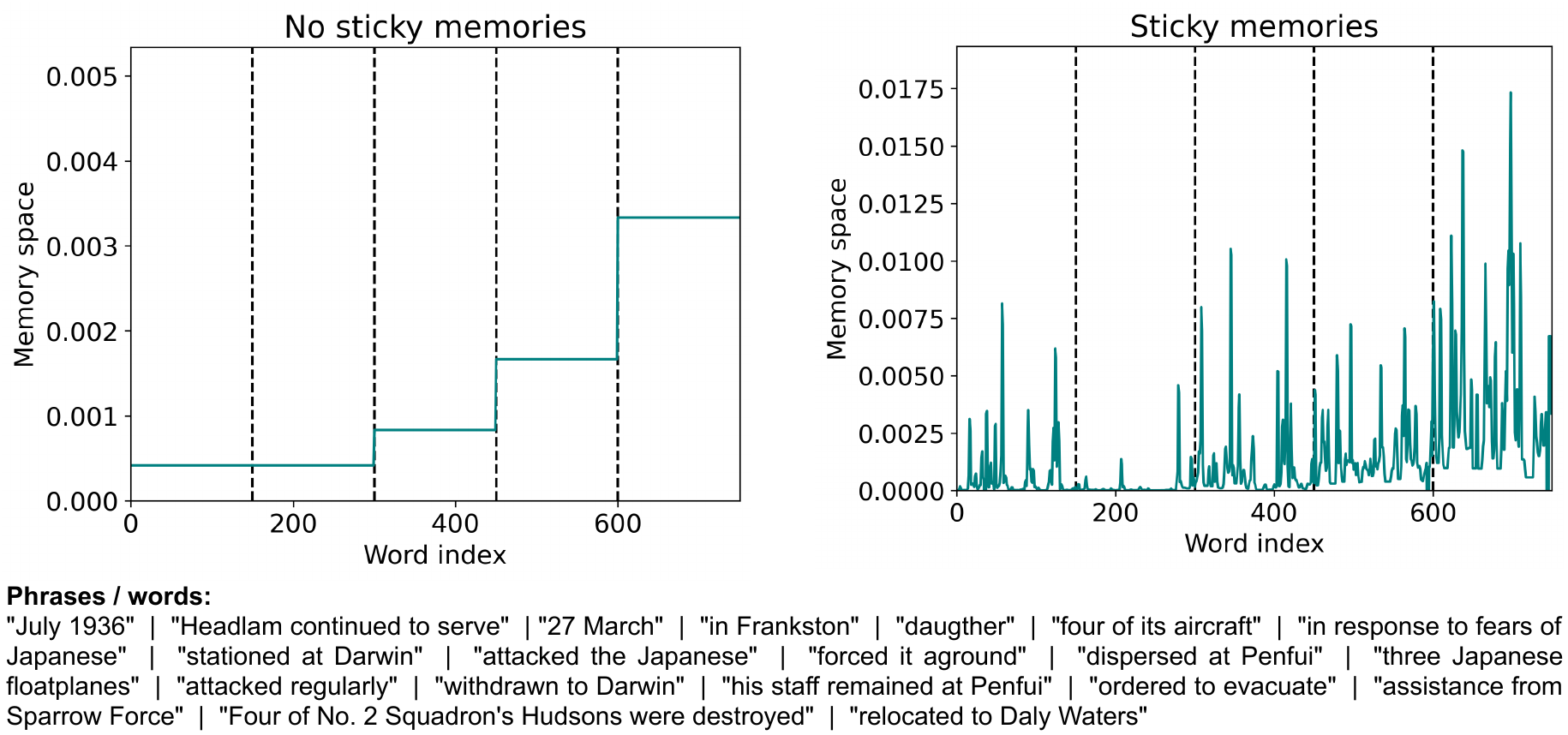}
  \caption{Example of the memory space attributed to each word in the last layer's long-term memory (after 5 updates) without / with the sticky memories procedure, along with the words / phrases which have the largest memory spaces when using sticky memories (top peaks with space$>.005$) Excerpt of the sequence being generated in this example: \textit{``Headlam became Officer Commanding North-Western Area in January 1946. Posted to Britain at the end of the year, he attended the Royal Air Force Staff College, Andover, and served with RAAF Overseas Headquarters, London.''} The dashed vertical lines represent the limits of the memory segments for the various memory updates.}
  \label{fig:example_sticky_mems_2}
\end{figure*}

To know whether the sticky memories approach attributes a larger space of the LTM's signal to relevant phrases or words, we plotted the memory space given to each word\footnote{The (Voronoi) memory space attributed to each word is half the distance from the previous word plus half the distance to the next word in the LTM's signal, being the word's location computed based on the sampled positions from which we evaluate the signal when receiving new memory vectors.} 
present in the long-term memory of the last layer when using and not using sticky memories. We present examples in Figs. \ref{fig:example_sticky_mems} and \ref{fig:example_sticky_mems_2} along with the phrases / words which receive the largest spaces when using sticky memories. We can see in these examples that this procedure does in fact attribute large spaces to old memories, creating memories that stick over time. We can also see that these memories appear to be relevant as shown by the words / phrases in the examples.


\section{Additional examples}
\label{sec:examples}
\begin{figure*}[t]
  \centering
    \includegraphics[width=\textwidth]{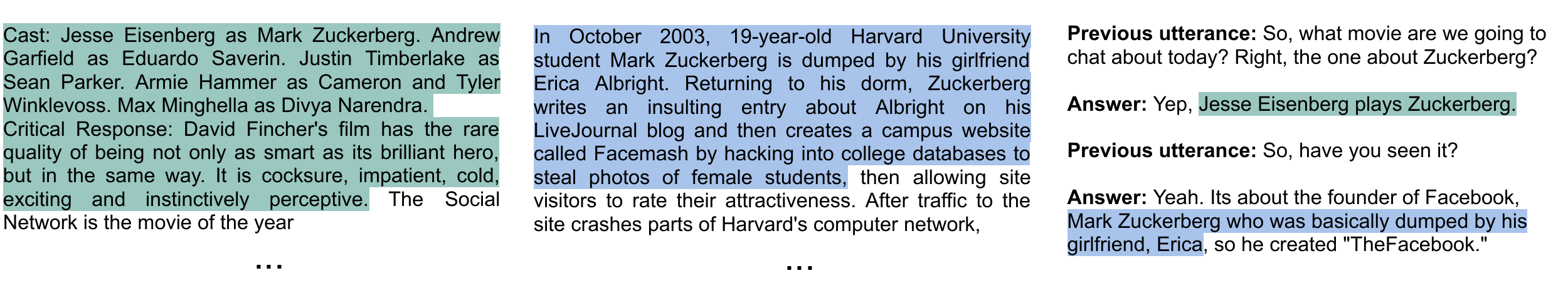}
  \caption{Examples of answers generated by \mbox{$\infty$-former} on a dialogue about the movie ``The Social Network''. The excerpts from the LTM that are more attended to throughout their generation are highlighted on each color correspondingly.}
  \label{fig:example_dgg2}
\end{figure*}

In Fig. \ref{fig:example_dgg2}, we show additional examples of utterances generated by \mbox{$\infty$-former} along with the excerpts from the LTM that receive higher attention throughout the utterances' generation.

Additionally, ground truth conversations concerning the movies ``Toy Story'' and ``La La Land'', for which the sticky memories are stated in Fig. \ref{fig:example_dgg_sm}, are shown in Tables \ref{tab:dialogue_toystory} and \ref{tab:dialogue_lalaland}, respectively.

\begin{table*}[h]
    \centering
    \small
    \begin{tabular}{p{\textwidth}}
    \toprule
        - Hi \\
        - Yo you really need to watch Toy Story. It has 100\% on Rotten Tomatoes! \\
        - Really! 100\% that's pretty good What's it about \\
        - It's an animated buddy-comedy where toys come to life \\
        - who stars in it \\
        - The main characters are voiced by Tom Hanks and Tim Allen \\
        - does it have any other critic ratings \\
        - Yep, metacritic gave it 95/100 and Cinemascore gave it an A \\
        - how old is it? \\
        - It's a 1995 film so 23 years \\
        - The old ones are always good :) I heard there were some sad parts in it is that true \\
        - Yeah actually, the movie starts off pretty sad as the toys fear that they might be replaced and that they have to move \\
        - is this a disney or dreamworks movie \\
        - Disney, pixar to be exact \\
        - Why do the toys think they will be replaced :( \\
        - they thought so because Andy was having a birthday party and might get new toys \\
        - What part does Tom Hanks play \\
        - Woody, the main character \\
        - How about Tim Allen \\
        - Buzz, the main antagonist at first then he becomes a friend \\
        - What kind of toy is Woody? \\
        - A cowboy doll \\
        - What is Buzz \\
        - A space ranger \\
        - so do all the toys talk \\
        - yep! but andy doesn't know that \\
        - Is andy a little kid or a teen \\
        - He's 6! \\
        - Sounds good. Thanks for the info. Have a great day \\
    \bottomrule
    \end{tabular}
    \caption{Ground truth conversation about movie ``Toy Story''.}
    \label{tab:dialogue_toystory}
\end{table*}

\begin{table*}[h]
    \centering
    \small
    \begin{tabular}{p{\textwidth}}
    \toprule
        - hey  \\
        - hey \\
        - i just watched la la land. It is a movie from 2016 starring ryan gosling and emma stone. they are too artists (one actress and one panist) and they fall in love and try to achieve their dreams. its a great movie \\
        - It's a wonderful movie and got a score of 92\% on rotten tomatoes \\
        - yes, i think it also won an oscar \\
        - Yes but I thought it was a little dull \\
        - metacritics rating is 93/100 as well its pretty critically acclaimed \\
        - the two leads singing and dancing weren't exceptional \\
        - i suppose it is not for everyone \\
        - It also sags badly in the middle  I like how Sebastian slipped into a passionate jazz despite warnings from the owner. \\
        - what do you think of the cover of "i ran so far away?"  in the movie, sebastian found the song an insult for a serious musician \\
        - I don't know, it is considered an insult for serious musicians not sure why \\
        - yeah \\
        - The idea of a one woman play was daring \\
        - it was interesting how sebastian joined a jazz fusion band he couldnt find real happiness in any of the bands he was in its hard  \\
        - It is considering she didn't know of any of that until she attended one of his concerts \\
        - yeah, that is daring  the movie kind of speaks to a lot of people.  she accussed him of abandoning his dreams  but sometimes thats what you have to do.  \\
        - Not nice that she leaves because he told her she liked him better when he was unsuccessful The play was a disaster so he didn't miss anything when he missed it. \\
        - yeah, but i dont blame her for dumping him for that  \\
        - She should didn't want to support him and she had to move back \\
        - id be pretty upset as well to boulder city nevada \\
        - yes she didn't want to forgive him, I didn't understand that \\
        - well because that was a big deal to her and he missed it \\
        - if she was with him when he was unsuccessful, she could have supported him to follow his dreams or other dreams \\
        - i suppose that is true \\
        - she wasn't successful either \\
        - yeah she wasnt nobody showed up to her play \\
        - so why the big hulabaloo about him \\
        - not sure \\
        - she was selfish I guess He missed her play because he had to go for a photo shoot with the band that he had previously missed \\
        - yeah but he should have kept better track and scheduled it better \\
        - this shows that he was trying to commit some and follow his dreams although not necessarily like them so she would be please if he didn't attend the photo shoot a second time, and came to her show \\
        - i definitely felt bad for both of them though in that scene \\
        - it's more of a do or don't he is still condemned I feel bad for him because he tried he tried to get her back by apologizing as well she didn't want any of it \\
        - yeah because she felt like he didnt care enough because he missed it he's the one that suggested the one woman play \\
        - They could have started all over again just like the beginning \\
        - maybe so \\
        - did she fail because of the one-woman play? she could have tried something else if she felt that \\
        - she wanted to give it a shot \\
        - she did and it failed, he did and it failed they just had to compromise so they could be together again, which  was how the happiness was  He signed up for the band after hearing her talking to her mom about how he is working\\
        - on his career I think he did a lot for her \\
    \bottomrule
    \end{tabular}
    \caption{Ground truth conversation about movie ``La La Land''.}
    \label{tab:dialogue_lalaland}
\end{table*}

\end{document}